\documentclass[acmsmall]{acmart}
\usepackage{subfig}
\usepackage[flushleft]{threeparttable}
\usepackage{color, colortbl}
\usepackage{multirow}
\usepackage{array}
\definecolor{Gray}{gray}{0.9}
\usepackage{soul}

\AtBeginDocument{%
  \providecommand\BibTeX{{%
    \normalfont B\kern-0.5em{\scshape i\kern-0.25em b}\kern-0.8em\TeX}}}

\setcopyright{acmcopyright}
\copyrightyear{2022}
\acmYear{2022}
\acmDOI{10.1145/1122445.1122456}

\begin{document}

\title{Understanding Postpartum Parents’ Experiences via Two Digital Platforms}

\author{Xuewen Yao}
\email{xuewen@utexas.edu}
\affiliation{%
  \institution{The University of Texas at Austin}
  \city{Austin}
  \state{Texas}
  \country{USA}
}

\author{Miriam Mikhelson}
\affiliation{%
  \institution{The University of Texas at Austin}
  \city{Austin}
  \state{Texas}
  \country{USA}
}
\email{mmikhelson@utexas.edu}

\author{Megan Micheletti}
\affiliation{%
  \institution{The University of Texas at Austin}
  \city{Austin}
  \state{Texas}
  \country{USA}
}
\email{m.micheletti@utexas.edu}

\author{Eunsol Choi}
\email{eunsol@utexas.edu}
\affiliation{%
  \institution{The University of Texas at Austin}
  \city{Austin}
  \state{Texas}
  \country{USA}
}

\author{S. Craig Watkins}
\email{craig.watkins@austin.utexas.edu}
\affiliation{%
  \institution{The University of Texas at Austin}
  \city{Austin}
  \state{Texas}
  \country{USA}
}

\author{Edison Thomaz}
\email{ethomaz@utexas.edu}
\affiliation{%
  \institution{The University of Texas at Austin}
  \city{Austin}
  \state{Texas}
  \country{USA}
}

\author{Kaya de Barbaro}
\email{kaya@austin.utexas.edu}
\affiliation{%
  \institution{The University of Texas at Austin}
  \city{Austin}
  \state{Texas}
  \country{USA}
}

\renewcommand{\shortauthors}{Yao et al.}

\begin{abstract}
 Digital platforms, including online forums and helplines, have emerged as avenues of support for caregivers suffering from postpartum mental health distress. Understanding support seekers' experiences as shared on these platforms could provide crucial insight into caregivers' needs during this vulnerable time. In the current work, we provide a descriptive analysis of the concerns, psychological states, and motivations shared by healthy and distressed postpartum support seekers on two digital platforms, a one-on-one digital helpline and a publicly available online forum. Using a combination of human annotations, dictionary models and unsupervised techniques, we find stark differences between the experiences of distressed and healthy mothers. Distressed mothers described interpersonal problems and a lack of support,  with 8.60\% - 14.56\% reporting severe symptoms including suicidal ideation. In contrast, the majority of healthy mothers described childcare issues, such as questions about breastfeeding or sleeping, and reported no severe mental health concerns. Across the two digital platforms, we found that distressed mothers shared similar content. However, the patterns of speech and affect shared by distressed mothers differed between the helpline vs. the online forum, suggesting the design of these platforms may shape meaningful measures of their support-seeking experiences. Our results provide new insight into the experiences of caregivers suffering from postpartum mental health distress. We conclude by discussing methodological considerations for understanding content shared by support seekers and design considerations for the next generation of support tools for postpartum parents.

\end{abstract}

\ccsdesc[500]{Human-centered computing~Empirical studies in collaborative and social computing}
\ccsdesc[300]{Applied computing~Psychology}




\keywords{Reddit, Digital Helpline, Postpartum Mood and Anxiety Disorders, Postpartum Depression, Clustering, LIWC, Text Analysis, Parenting, Social Media, Natural Language Processing}

\maketitle

\section{Introduction}
Postpartum mood and anxiety disorders (PPMADs) affect up to one in five women globally \cite{post_partum_dep}, with 85\% of postpartum women reporting some form of mood disturbance \cite{henshaw2003}. Maternal PPMADs are associated with many adverse effects on child development across cognitive, motor, and mental health domains \citep{murray_et_al_1996), cumming_etal_1994, goodman_gotlib_1999, gotlib_lee_1996, weissman_etal_2006}. Given the ubiquity of PPMADs and their known consequences for both mothers and children, it is essential to thoroughly understand the challenges and needs of caregivers with postpartum distress. 

Nearly 60\% of mothers with PPMADs symptoms are not professionally diagnosed, and 50\% of diagnosed mothers are left untreated, with poor women and women of color being disproportionately impacted \cite{howell_etal_2005}. However, new parents are increasingly turning to peer-to-peer digital platforms to share their experiences and access postpartum support. Compared with traditional face-to-face therapies and their extension to video-conferencing platforms, digital platforms represent a convenient outlet for parents burdened with constraints on time, money or geography \cite{online_social_support_2001}. Therefore, the aim of this paper is to better understand the experiences of postpartum support seekers using digital platforms, with the ultimate goal of developing better postpartum support bots.


Knowledge about the specific concerns facing caregivers experiencing PPMADs could help to design better tools to support them.  Psychoeducation is a common component of many interventions for PPMADs \cite{cuijpers2008psychological}, but there is mixed evidence for its effectiveness in reducing psychological distress in the postpartum period \cite{austin2005clinical,honey2002brief}. One possible explanation for this is that some educational materials may be mismatched to the specific concerns of individuals experiencing postpartum distress \cite{universal_psychoed_intervention_review}, and in turn not provide the intended support. Given that increasing numbers of caregivers are turning to digital platforms for support, analyses of shared content can provide rich insights into their concerns \cite{whatsapp, diyi_onlineMedia}. However, we know of no prior work that has examined support seekers' content on digital platforms as a source of information about the concerns of caregivers experiencing symptoms of PPMADs.


As such, in the current study we examine content shared on two digital platforms dedicated to supporting caregivers with postpartum distress. The first platform is a text-message based "helpline" platform, developed by the Postpartum Support International (PSI) \cite{psi}. PSI is the largest  international non-profit organization serving the needs of pregnant, post-loss, and postpartum support seekers. In a collaboration with PSI, our goal was to analyze de-identified content from hotline messages, with the ultimate goal of developing a chatbot that could provide educational material and support around the most common concerns shared by women experiencing postpartum distress. To assess the extent to which the concerns shared on the PSI helpline generalized to the broader community of women experiencing postpartum distress, we also examined posts shared by caregivers seeking support on two Reddit communities, one geared toward caregivers with postpartum mental health concerns and one geared toward general support for new parents. 

The PSI helpline and Reddit differ on many dimensions and thus may provide very different opportunities for sharing - and support. Most broadly, the goal of chat-based platforms, including digital helplines, is to provide targeted emotional support and information to support seekers through one-on-one exchanges. By contrast, online forums such as Reddit provide support seekers with information or discussion via crowdsourced peer support. Such platforms are typically public, meaning support seekers can reach a large audience. However, the availability and quality of support is not guaranteed or monitored, with one study finding that over 30\% of Reddit posts received no comments at all \cite{reddit_conversation_patterns}.


Another key difference involves the individuals providing the support. Digital helplines are usually maintained by organizations that train volunteers or paid staff to provide professional counselling. By contrast, in online forums, an anonymous group of peers provides support. Perceived differences in the expertise and anonymity of supporters could also impact the content shared between digital helplines and online forums. For example, support seekers who feel stigmatized are more comfortable sharing on an anonymous online forum than speaking with a professional \cite{health_forums,virtual_voices}. As such, examining support seekers' concerns across these platforms can provide us a more robust understanding of the needs of caregivers experiencing postpartum distress.

We compared content from caregivers seeking support for symptoms of PPMADs, henceforth "caregivers with postpartum distress", with caregivers who were new parents in the postpartum period who were not specifically seeking support for postpartum distress. This allowed us to determine whether potential differences we observed between the Reddit PPMADs forums and the PSI helpline were due to differences between the platforms rather than differences in distressed support seekers' concerns, per se. Finally, while mothers continue to do the majority of childcare globally \cite{mother_majority_childcare}, it is increasingly recognized that fathers and other caregivers of infants also suffer from PPMADs, with rates of paternal depression being as high as 10.4\% in some studies \cite{paternal_depression}. Given cultural attitudes and gender norms, new fathers are likely to have distinct postpartum experiences from those shared by mothers. Thus, we also include in our investigation the experiences of fathers and other non-maternal caregivers sharing to the digital parenting platforms.

Comparing content from  "distressed" support seekers sharing on the helpline and postpartum mental health subreddits to content from ostensibly "healthy" support seekers posting on a general parenting subreddit, our paper addresses three main empirical questions:

\begin{itemize}
    \item \textbf{What are the concerns of postpartum support seekers?}  Are the challenges shared by support seekers on postpartum mental health platforms different from those shared on general new parent platforms, or are the types of concerns similar but more intense?  
    \item \textbf{What are the psychological states of postpartum support seekers sharing on these digital platforms?} Does this differ according to whether they share on a postpartum mental health platform vs. a general parenting platform? Are there differences in the expressed affect shared on online forums or on chat-based platforms? 
    \item \textbf{What are the goals of postpartum support seekers?} Are support seekers looking for validation, advice, or concrete resources? And is this systematically related to their choice of platform?

\end{itemize}

We use a combination of human annotations, existing dictionary-based text analysis, and unsupervised clustering techniques to analyze these data to provide complementary insights into our questions. Thus, our final question was focused on methodology: 
\begin{itemize}

\item \textbf{What are the contributions of machine analyses versus human annotations in understanding support seekers' concerns?} We compared our findings from dictionary-based methods, unsupervised clustering and human annotation to assess whether they provide unique or  complementary insights into support seekers' needs.

\end{itemize}

\section{Related Works}
There is a sizable body of literature examining content shared by digital support seekers in both parenting and mental health communities.

Some of this work has been used to predict mental health status using both private and publicly available posts. For example, De Choudhury et al. utilized Twitter data to analyze the posting patterns and language usage before and after birth by new mothers and built models to predict significant postpartum changes in mothers \citep{munmun_twitter2, predictingtwitter, fbpredict}. Additionally, Shatte et al. used Reddit to identify fathers at risk of postpartum depression \cite{fathers}. These prior efforts leverage language and dictionary-based features to provide insight into the states of women experiencing postpartum depression. For example, work from De Choudhury and others has found that women experiencing symptoms of postpartum depression, use more 1\textsuperscript{st} person singular pronouns ("I") and fewer 2\textsuperscript{nd}, 3\textsuperscript{rd} person pronouns ("you", "he", "she"), fewer articles ("the") and more past tense \citep{predictingtwitter, fbpredict, depression_pasttense}. They also use more negative emotion words, fewer positive emotions words, and more negative and informal speech \citep{fbpredict, negative_cognitive_styles}. However, while these differences in grammar and word use indicate that postpartum support seekers can be distinguished from healthy controls, they provide only a basic description of the challenges faced by postpartum support seekers. For example, they do not allow us to know what kinds of support would be most helpful to provide.

A much smaller body of literature has focused on rich characterizations of the \textit{content} of experiences shared by digital support seekers with human insights and annotations.  For example, a recent work by Yadav et al. \cite{whatsapp} looked into conversations from WhatsApp-mediated online health support groups for expectant and new mothers in rural India. By qualitatively analyzing and annotating chat records among 588 group members, they found that the most frequent form of support sought and provided was informational support. Emotional support, though existing, was quite limited potentially due to moderation role and lack of mutuality in fostering experience sharing. Another work by Gao et al. \cite{diyi_onlineMedia} identified and collected data from 46 parenting subreddits to understand parenting topics from infancy to preschool. Mining 602K posts and 10M comments, this work detailed over 30 topics discussed frequently by parents, and demonstrated that across parents' history of posting, discussed topics shifted over time, in line with children's growth. To parse this large volume of text, Gao et al. leveraged supervised models trained with ground-truth data from a complex, iteratively developed annotation scheme. A key benefit of such a scheme is that it can be designed to reflect the particular questions of greatest interest to the researcher. Additionally, a human-generated annotation scheme can provide very rich insight into the support seekers posts relative to more automated approaches. 

Other recent papers have examined the content of support seekers' digital posts using unsupervised techniques. For example, one paper examined how concerns of mental health support seekers changed before and during the COVID-19 pandemic \cite{reddit_nlp_covid}. Using topic modeling and unsupervised clustering, they discovered that during the pandemic there were twice as many posts regarding suicidality and loneliness, and the clusters surrounding self-harm and entertainment emerged. These papers combine clustering techniques with both quantitative and qualitative information to provide insights into support seekers' concerns. Such "bottom up" methods start with the data rather than the researcher's pre-conceived notions of relevant concerns, and thus may identify unexpected underlying patterns within the data. However, the categories discovered may not be of clinical relevance, for example, sentences likely to contain numbers or dates. 
Finally, the labelling of clusters ultimately depends on human researchers examining these groups with little insight into how they were generated. While systematic techniques can be used to determine cluster identity, misinterpretations can still occur, especially in the case of outliers.

\section{Analysis plan} 



Given the complementary strengths and weaknesses of supervised and unsupervised text-analysis techniques, we used a combination of methods to gain insight into the experiences of postpartum support seekers. First, we iteratively developed a rich annotation scheme to obtain highly reliable human-generated insights on a subset of data. Additionally, we leveraged the dictionary-based linguistic text analysis software LIWC \cite{liwc} to analyze our conversations and posts at scale. Given the existing literature utilizing such tools, we used them to provide a basic validation of our other approaches and to allow comparison with prior work. Finally, we use clustering techniques to obtain similarities and differences within and between datasets from different communities.  

In particular, recent developments in natural language processing have resulted in a new class of enormous pre-trained language models \cite{pretrain_survey}, such as GPT-3 \cite{gpt3} and BERT \cite{bert}. These pre-trained models are parallelizable and good at obtaining contextual relationships across long texts and they have already proven their excellency when adapted to a number of complicated downstream tasks \cite{nlp_pretrained} such as machine translation, question answering, and natural language inference. Thus, to understand the content from support seekers we leveraged representations extracted by BERT, a transformer-based machine learning technique, and performed clustering on the datasets.

\section{Datasets} \label{sec:dataset}
The primary contribution of this work is to leverage digital platforms to understand the experiences of caregivers suffering from postpartum distress and how they differ from those of healthy parents. Additionally, we evaluate potential differences in support seekers shared content across different digital platforms. To do so, we collected, annotated and analyzed text data from postpartum support seekers as well as healthy parents from two different digital platforms 1) a one-on-one synchronous helpline dataset maintained by an international postpartum support organization, Postpartum Support International (PSI), and 2) an asynchronous dataset of threads collected from Reddit users posting on postpartum mental health pages or more generic new parent pages. As such, this dataset represents the best-in-practice in accessible, professional digital support specific to PPMADs compared to alternatives, such as 7 Cups \cite{7cups} and Crisis Text Line \cite{crisistextline}. Reddit, on the other hand, is favoured over other forum-based or asynchronous platforms such as TalkLife \cite{talklife} and Facebook \cite{facebook} because of its popularity and public accessibility \cite{why_reddit}. Additionally, its communities are topic-based, allowing us to easily identify and study our targeted population, new parents and caregivers suffering from postpartum distress \cite{why_reddit}. 

Due to the sensitive nature of our datasets and study, we have taken multiple measures to ensure the privacy of the support seekers whose content we used in this study. We worked closely with PSI on this project from visioning to analysis to ensure that the data was used in ways that would respect and benefit the PSI community. Our data transfer agreement ensured that all data would be de-identified prior to analyses, that no efforts would be made to re-identify support seekers, and that the data would not be shared beyond our specifically named shared team. Our research was approved by the Institutional Review Board (IRB) at [name redacted for anonimity] and all researchers working on this project finished IRB training to ensure proper behaviors when handling data. Additionally, to further enhance the privacy of the support seekers while  preserving the rich descriptive character of the original content, we have paraphrased all quotes and edited or "mixed and matched" relevant details across multiple participants, following prior works \cite{diyi_onlineMedia}. As such, there is no direct relationship between the printed "quotes" and the material shared directly by PSI or Reddit support seekers.

The statistics of our collected datasets are shown in Table~\ref{tab:datasets}, along with information regarding the filtered and annotated data (detailed below). On average, support seekers wrote 12.81 sentences in a helpline conversation, versus 16.61 and 9.69 sentences when they posted on r/ppd and r/NewParents. 



\subsection{Helpline Data} \label{sec:helpline}
As part of an extended collaboration with PSI, we were granted access to de-identified content shared by support seekers on the PSI helpline. The PSI helpline is a 24/7 helpline where trained volunteers offer support seekers emotional validation and resources. With a signed information transfer agreement, we acquired access to all helpline text conversations in their database. 
We collected a total of 7014 text conversations between 1/26/2019 – 10/12/2020. They contained 65062 individual messages, 29108 (44.8\%) of which were sent from support seekers. Prior to any analyses of content, all conversations were de-identified by finding personally identifiable information with named-entity recognition \cite{spacy} and replacing them with placeholders such as PSI\_PERSON and PSI\_PLACE. Although geographical location was de-identified, the helpline provided support in every state of the US, Canada, and wherever PSI Support Coordinators were located internationally \cite{psi_2022}. 99.6\% of helpline conversations were conducted in English.

\subsection{Reddit Data}
We collected online forum content from three subreddits. The first two are r/Postpartum\_Depression (4.7K users), r/postpartumdepression (1.6K users), both focused on users who identify as having postpartum mental health distress. The third subreddit is r/NewParents (173K users), a general forum for parents in the first year of caregiving (i.e.  not specific to postpartum mental health issues), where ostensibly "healthy" support seekers post. All Reddit data was collected using Reddit API PRAW.\footnote{We access the data on 10/25/2021 with \url{https://github.com/praw-dev/praw}} All posts were in English. Although geographical information was not publicly available for every user, we expected locations to include the US (where approximately 50\% of Reddit users reside \cite{clement_2022}) and almost all countries. We initially gathered 1124 posts from r/Postpartum\_Depression and r/postpartumdepression (these posts are combined in further analysis and referred to as r/ppd going forward) and 950 posts from r/NewParents. We manually filtered out any posts that did not contain concerns, including posts sharing exclusively positive information, surveys, public announcements, resources sharing, and incomplete posts (either too short to infer useful information or the main content was an image/GIF which cannot be scraped). After removal, 923 posts from r/ppd and 700 posts from r/NewParents remained.

\begin{table*}
  \caption{Datasets}
  \label{tab:datasets}
  \begin{tabular}{cccc}
    \toprule
    \textbf{\# of samples} & \textbf{Helpline} & \textbf{r/ppd} & \textbf{r/NewParents} \\
    \midrule
    Collected & 7014 & 1124 & 950 \\
    Filtering & 4835 & 923 & 700 \\
    Annotated & 730 & 120 & 120\\
    \bottomrule
\end{tabular}
\end{table*}


\section{Methods}
To provide complementary insights from both a human and data perspective, we leveraged three methods to analyze our collected data: manual annotation, automated linguistic text analysis with the LIWC lexicon, and unsupervised learning using sentence representation clustering. 


\subsection{Manual Annotation}

\paragraph{Data} We selected a random subset of 730 helpline conversations (10.41\%) and 240 posts from Reddit (120 from r/ppd, 13\%, and 120 from r/NewParents, 17.14\%) for annotation by the trained research assistants. During our annotation of helpline data, we found that 58 conversations were surveys sent from PSI, which we excluded from further analysis. Out of the remaining 672 conversations, 407 conversations (60.6\%) were from mothers in the postpartum period looking for support. The rest of the conversations included: requests from pregnant women or women who lost their child (51, 7.6\%), requests from fathers (17, 2.5\%), requests from other family members (6, 0.89\%), requests from friends (4, 0.60\%), and requests from other individuals including social workers or mental health providers inquiring about the hotline (15, 2.23\%). The remaining 230 conversations (34.2\%) were inconclusive, meaning that annotators could not infer the role of the support seeker nor any details about their experiences. In most cases, the support seeker initialized the conversation with a greeting but didn't continue the exchange. Given the relatively small number of posts identified from fathers and other non-maternal caregivers, we exclusively conducted our dictionary-based text analysis and clustering analyses on posts from mothers seeking support. However, we annotated all identified conversations from fathers, other caregivers and individuals proximate to caregivers, the results of which we share in Section~\ref{sec: others}. To further ensure the quality of our dataset, conversations with fewer than 3 conversational turns and 50 words were excluded from clustering and text analyses, resulting in 4835 conversations used for analyses. No such thing was done for Reddit data as we manually filtered out inconclusive posts during data collection. We found for r/NewParents and r/ppd, 116 and 102 posts were authored by mothers, and with 4 and 13 posts authored by fathers, respectively. The details of our datasets can be found in Table~\ref{tab:datasets}.

\begin{table*}
  \caption{Annotation Scheme of Concerns: Definition, Inter-rater Agreement (Kappa) and the Number of Conversations/Threads Containing the Concern}
  \label{tab:annotation}
  \begin{tabular}{lp{8.2cm}lr}
    \toprule
    \textbf{Concern} & \textbf{Definition} & \textbf{Kappa} & \textbf{\#}\\
  	\midrule
    \multicolumn{2}{l}{\textbf{Interpersonal Problems}}   &  &  \\
    \hline
    \rowcolor{Gray}
    Violence & Violent threat from or interaction with another person & 0.657 & 3\\
    Partner & Conflict or disagreement (non-violent) with partner & 0.785 & 99\\
    \rowcolor{Gray}
    Family & Conflict or disagreement (non-violent) with family member & 0.495 & 45\\
    \hline
    \multicolumn{2}{l}{\textbf{Childcare Concerns}} & &\\
    \hline
    \rowcolor{Gray}
    Breastfeeding & Breastfeeding, nursing, and formula-feeding & 0.503 & 55\\
    Crying & Child fussing or crying & 0.465 & 38\\
    \rowcolor{Gray}
    Sleep & Child's sleep & 0.576 & 43\\
    Health & Medical concerns & 0.560 & 25 \\
    \rowcolor{Gray}
    Errands & Overwhelming amount of tasks related to child care & 0.5 & 11 \\
    Products & Asking for baby product and/or brand recommendations & 1 & 8 \\
    \rowcolor{Gray}
    Non-milk feeding & Feeding child non-milk food-- quantity, frequency, etc. & 0.478 & 7\\  
    Behaviors & Questions regarding milestones or other child behaviors & 0.561 & 20 \\
    \hline
    \multicolumn{2}{l}{\textbf{Life Stressors}}  & &\\
    \hline
    \rowcolor{Gray}
    COVID-19 & Problems related to the COVID-19, including isolation & 0.666 & 43 \\
    Death & Death of a loved one & 1 & 8\\
    \rowcolor{Gray}
    Finances & Concerns about having enough money, not being able to afford health care/insurance & 0.701 & 21\\
    Employment & Problems with employment, going back to work after leave & 0.441 & 14\\
    \rowcolor{Gray}
    Medical & Personal health concern, severe and/or requiring medical attention & 0.666 & 7\\
    Physical discomfort & Personal physical pain, mild and not requiring medical attention such as a headache & 0.5 & 26\\
    \rowcolor{Gray}
    Medication & Concerns about taking medications, often antidepressants &  0.272 & 40 \\
    Insomnia & Issues with lack of sleep & 0.264 & 52\\
    \rowcolor{Gray}
    Errands & Overwhelming amount of tasks overall, including child care & 1 & 10\\
    \hline
    \multicolumn{2}{l}{\textbf{Transition to Motherhood}} & &\\
    \hline
    \rowcolor{Gray}
    Lifestyle pre-birth & Missing autonomy and lifestyle before having a child & 0.833 & 17\\
    Body image & Insecurity about body image or weight & 0.699 & 14\\ 
    \rowcolor{Gray}
    At home with child & Adjusting to being at home with child &  0.665 & 8\\
    
    No time to self & No time to take a break or for time alone & 0.703 & 37\\
    \rowcolor{Gray}
    Bonding with baby & Trouble with forming connection with child & 0.558 & 22 \\
    Commitment & Concerns about or processing the long-term commitment of having a child & 0.833 & 10\\
    \rowcolor{Gray}
    Parenting confidence & Doubting parenting abilities, comparisons to other parents & 0.489 & 73 \\
    Prenatal trauma & Reflecting on prenatal and/or child-birth difficulties & 0.822 & 40\\
    \hline
    \multicolumn{2}{l}{\textbf{Lack of Support}} & &  \\
    \hline
    \rowcolor{Gray}
    Personal & Lack of social support, feeling misunderstood, not having someone to confide in & 0.652 & 112 \\
    Professional & Lack of support or inadequate support from health care providers, lactation consultants, etc. & 0.705 & 68 \\
  \bottomrule
\end{tabular}
\end{table*}

\paragraph{Annotation Scheme}
We developed the annotation scheme with the goal of comprehensively categorizing support seekers' concerns, their reported psychological symptoms and the types of requested support. After reading and discussing the prevailing themes of concerns found in 100 of the longest conversations from the helpline data set, five research assistants, including two of the authors drafted an initial annotation scheme with items divided into categories associated with support-seeker concerns, psychological states, and goals. Support-seeker concerns included five subcategories, i.e. interpersonal problems, childcare concerns, life stressors, transition to motherhood, and lack of support.


\begin{table*}
  \caption{Annotation Scheme of Psychological States: Definition, Inter-rater Agreement (Kappa) and the Number of Conversations/Threads Containing the Psychological State}
  \label{tab:annotation2}
  \begin{tabular}{p{3.2cm}p{7.7cm}cr}
    \toprule
    \textbf{Psychological States} & \textbf{Definition} & \textbf{Kappa} & \textbf{\#} \\
    \midrule
    \rowcolor{Gray}
    Depressive Mood & Emotions and thoughts associated with clinical depression, sub-clinical sadness, hopelessness, and guilt & 0.533 & 363 \\
    Anxiety & Emotions and thoughts associated with clinical anxiety or panic, sub-clinical worry, fear, and irritation & 0.713 & 225\\
    \rowcolor{Gray}
    Unhealthy coping behavior & Harmful behavior intended to manage distress, such as excessive alcohol consumption or food restriction & 0.283 & 12 \\
    Severe Symptom & Suicidal thoughts, self-harm, or other indication of potential safety concern & 0.721 & 50\\
    \bottomrule
\end{tabular}
\end{table*}

\begin{table*}
  \caption{Annotation Scheme of Goals: Definition, Inter-rater Agreement (Kappa) and the Number of Conversations/Threads Containing the Goal}
  \label{tab:annotation_specific_request}
  \begin{tabular}{p{3.2cm}p{7.7cm}cr}
     \toprule
     \textbf{Goals}& \textbf{Definition} & \textbf{Kappa}  & \textbf{\#}\\ 
     \midrule
     \rowcolor{Gray}
     Local Coordinator & Connecting with a PSI local coordinator & 0.498 & 21\\ 
     Talk With Someone & Wanting to talk or vent to someone right now & 0.644 & 154\\
     \rowcolor{Gray}
     Health Care Provider & Looking for a therapist, lactation specialist, or other specialized provider & 0.414 & 60\\
     General Support & Asking for help, including specific advice to non-specific requests for support & 0.611 & 341\\
     \rowcolor{Gray}
     PPD Information & Questions about postpartum depression symptoms & 0.368 & 46\\
     Resource Barrier & Looking for support given a limitation such as no insurance or no specialists in the area  & 0.563 & 32\\
     \rowcolor{Gray}
     PSI Information & Questions about services provided by PSI & 0.740 & 64\\
\bottomrule
\end{tabular}
\end{table*}

During preliminary rounds of annotation, research assistants began annotating randomly selected helpline conversations to update and collaboratively evaluate the comprehensiveness of topics included in each category. The annotation scheme was thus dynamically updated with additional topics identified by the research assistants until there was consensus on a finalized annotation scheme. Additionally, due to the similarities in content between interpersonal concerns and lack of support, we later merged these two categories into a single category, Interpersonal Problems, for analyses.  The finalized coding scheme is available at Table~\ref{tab:annotation}, Table~\ref{tab:annotation2}, and Table~\ref{tab:annotation_specific_request}. Annotators used doccano \cite{doccano} software to code each sentence of each selected conversation and post with all applicable concerns and states of mothers. Goals were coded at the level of the conversation or post. Individual concerns, states, and goals were not considered mutually exclusive, that is, each sentence or each post/conversation could be coded with multiple annotations from within or between these categories. Individual concerns which occur less than 5 times in the data set were removed from further analyses. 

Inter-annotator agreement (Cohen's Kappa~\cite{kappa}) was evaluated for a subset of the annotations. We calculated agreement at the level of the conversational turn for the helpline conversations and at the level of the post for Reddit data as the content or context might change as a conversation develops but rarely for a single post, and the transition captured can be useful for future research. For example, if in a conversation turn, both raters have annotated breastfeeding concerns, then this is considered a match even if one rater has annotated it once and the other has annotated it twice. For the category of concern, raters obtained a mean kappa of 0.635, indicating moderate agreement among five raters \cite{why_0.4}. Most individual concerns, such as interpersonal problems: partner, childcare concerns: sleep, have a kappa between 0.4 and 0.85, indicating fair, good, or excellent agreement between raters. A few concerns, such as life stressors: medication and insomnia, have a kappa lower than 0.4, indicating we don't have agreement for those concerns. We discarded those individual concerns, psychological states (Table~\ref{tab:annotation2}), and goals (Table~\ref{tab:annotation_specific_request}) with a kappa score lower than 0.4 and they are not involved in our analysis. 

\subsection{Linguistic Text Analysis}

We use dictionary-based text analysis tool, LIWC, to provide markers of well-established affect-related words, topics, and language patterns at large scale. Drawing from past research, we examined the following categories: personal pronouns, articles, negation, affect words, time orientations, informal speech, and personal concerns. 

Given that the word categories provided by LIWC could ostensibly capture aspects of support-seeker affect and personal concerns but not goals, we excluded sentences in appreciation and positive sentiment and coordination clusters (see Section~\ref{sec:cluster}) to focus on the parts of the conversations where support seekers described their situations, feelings, questions, and needs. LIWC outputs the percentage of words in each individual category for each conversation/post, after which we used a student's t-test to compare the differences between the helpline and the two Reddit datasets.

\subsection{Clustering} \label{sec:cluster}
We aim to use clustering to find patterns in datasets overlooked by manual annotation as human coding can be subjective and clustering is data-oriented. Specifically, we would like to find similarities and differences in semantic information within and between datasets. To achieve that, we applied K-means clustering on all sentences extracted from the 3 datasets.



\begin{table*}
  \caption{Clusters with Definitions and Examples}
  \label{tab:clusters}
  \begin{tabular}{p{1.8cm}p{5.4cm}p{5.4cm}}
    \toprule
    \textbf{Name} & \textbf{Cluster Definition} & \textbf{Examples} \\
    \midrule
    \multirow{2}{1.6cm}{Neutral Information (18.86\%)} & a "number" cluster about "how long", "how old", and "since when" & \textit{My baby is 2 months old.}\\
    \cline{2-3}
    & neutral descriptions or background information &   \textit{I'm a stay at home mom.}\\
    \hline
    \multirow{2}{1.6cm}{Negative Situations (28.57\%)}  & describing problems with negative feelings  (containing many nouns and adjectives) & \textit{I'm busting my ass day in and day out and my husband doesn't even notice that I'm falling apart.}\\
    \cline{2-3}
    & a "number" cluster containing detailed descriptions of a negative situation & \textit{It's 4 am and I've been trying to soothe my baby for at least 3 hours now.}\\
    \hline
    Lack of Control (8.49\%) & a "no" cluster about things they don't know, couldn't do or don't have  & \textit{I don't know if it's postpartum or just the baby blues.}\\
    \hline
    \multirow{2}{1.6cm}{Depressing Feelings (13.73\%)} & sharing of negative or depressing feelings & \textit{I feel like such a burden to everyone around me.}\\
    \cline{2-3}
    \textbf{} & negative feelings partially containing "body" or "health" words & \textit{I always get horrible migraines when I'm this anxious.}\\
    \hline
    Questions (1.80\%) & questions or needs on resources, emotional validation, the function of the helpline, and so on &  \textit{Do you have any tips on how I can stop feeling so worried all the time?}\\
    \hline
    Appreciation and Positive sentiment (11.36\%) & appreciation for the support and positive feelings & \textit{I really appreciate all the resources and tips.} \& \textit{Thank you for listening and for the bedtime routine ideas, that feels like a good start.}\\
    
    \hline
    \multirow{2}{1.6cm}{Coordination (17.19\%)} & Reaching out and coordination on support   & \textit{I haven't gotten a call back from my local coordinator yet.}\\
    \cline{2-3}
    & short sentences presenting their name and/or location & \textit{Sure, my name is PSI\_NAME and I live in PSI\_PLACE.}\\
    
    \bottomrule
\end{tabular}
\end{table*}

We performed a sentence-level clustering algorithm on all sentences from support seekers. Inspired by the work of \cite{shao_2020} on contextual topic identification, we extracted sentence embeddings using both Latent Dirichlet Allocation (LDA) and BERT. We then clustered the concatenation of these two vectors using K-means and subsequently performed hyperparameter tuning to identify the best set of parameters (i.e. number of clusters) based on their coherence and silhouette scores. We identified 11 clusters from all sentences by support seekers, and after analyzing the top 20 key words extracted using TF-IDF and 50 random sentences from each cluster, we manually grouped them into 7 clusters. They are summarized in Table~\ref{tab:clusters}. 
	
\section{Results}
This section leverages annotations, linguistic text analysis, and clustering results to investigate the concerns, psychological states, and goals of postpartum support seekers across two different digital platforms. Additionally, we investigated the contributions of machine analyses vs. human annotations in understanding support seekers’ concerns. 

\subsection{What are the concerns of postpartum support seekers?} \label{sec: topics}
\begin{figure}%
    \centering
    \subfloat[ Individual concerns of helpline  ]{{\includegraphics[width=0.5\linewidth]{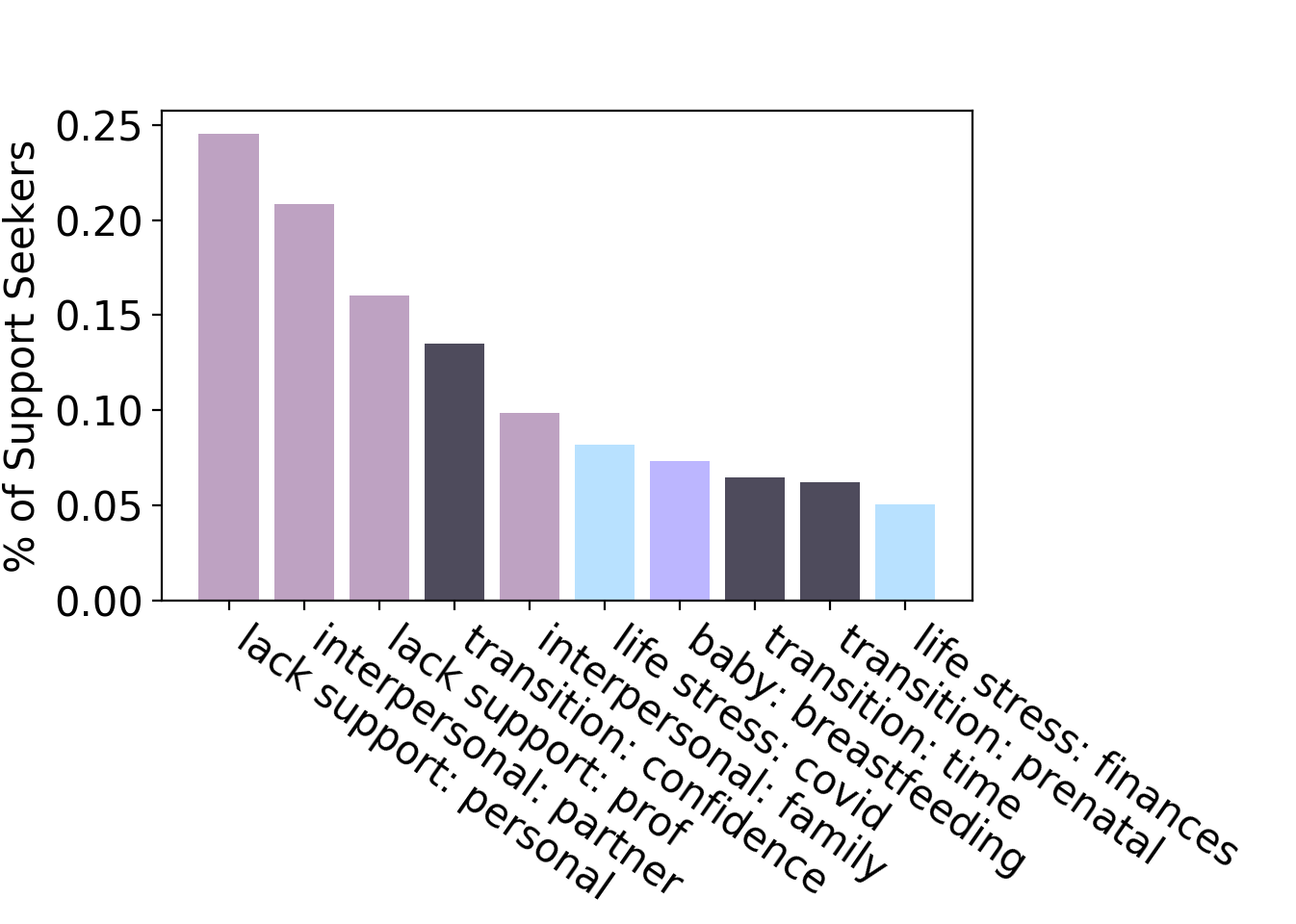} }}%
    \subfloat[\centering Concern categories of helpline]{{\includegraphics[width=0.5\linewidth]{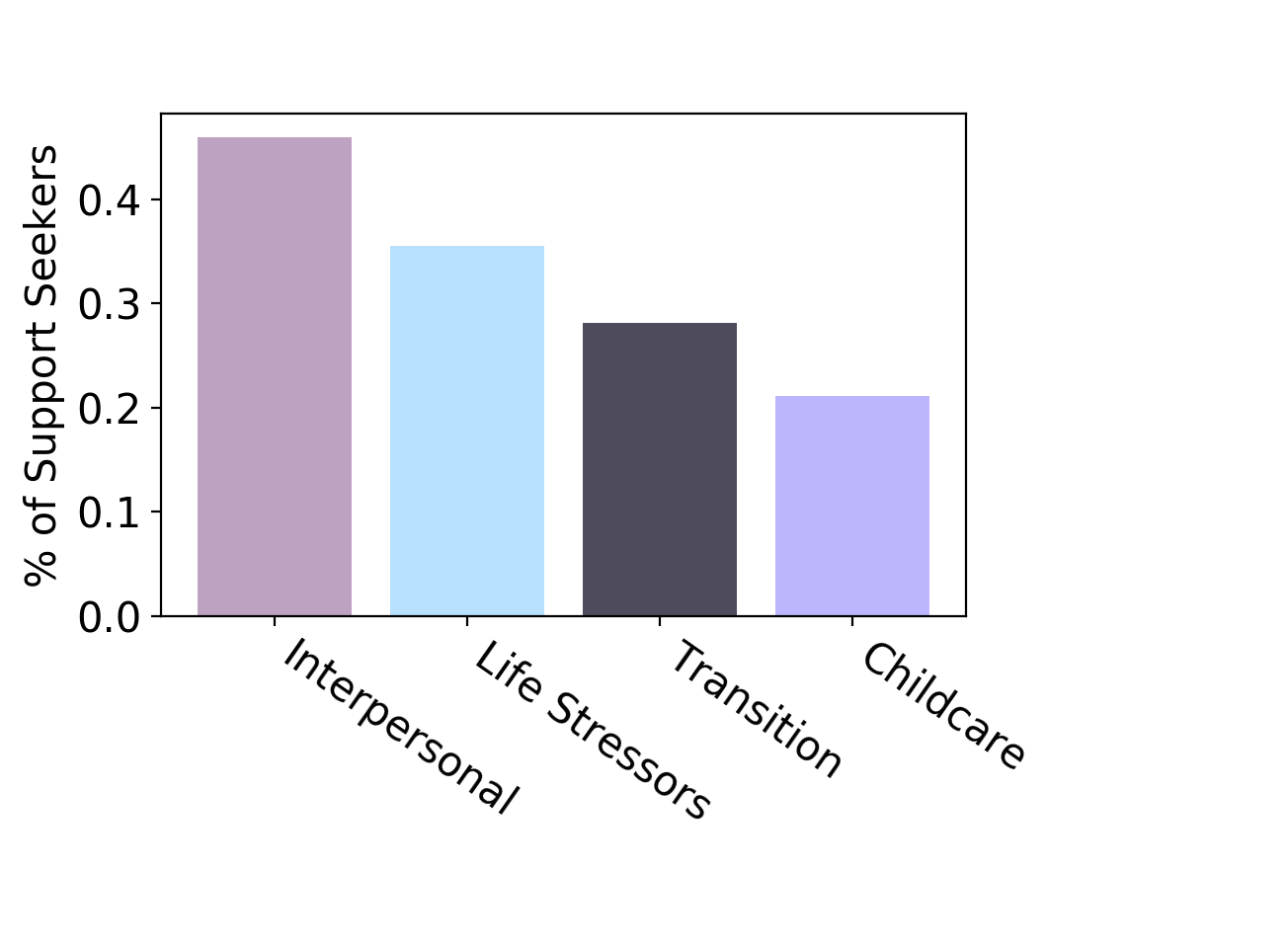} }} \\
    \subfloat[\centering Individual concerns of r/ppd]{{\includegraphics[width=0.5\linewidth]{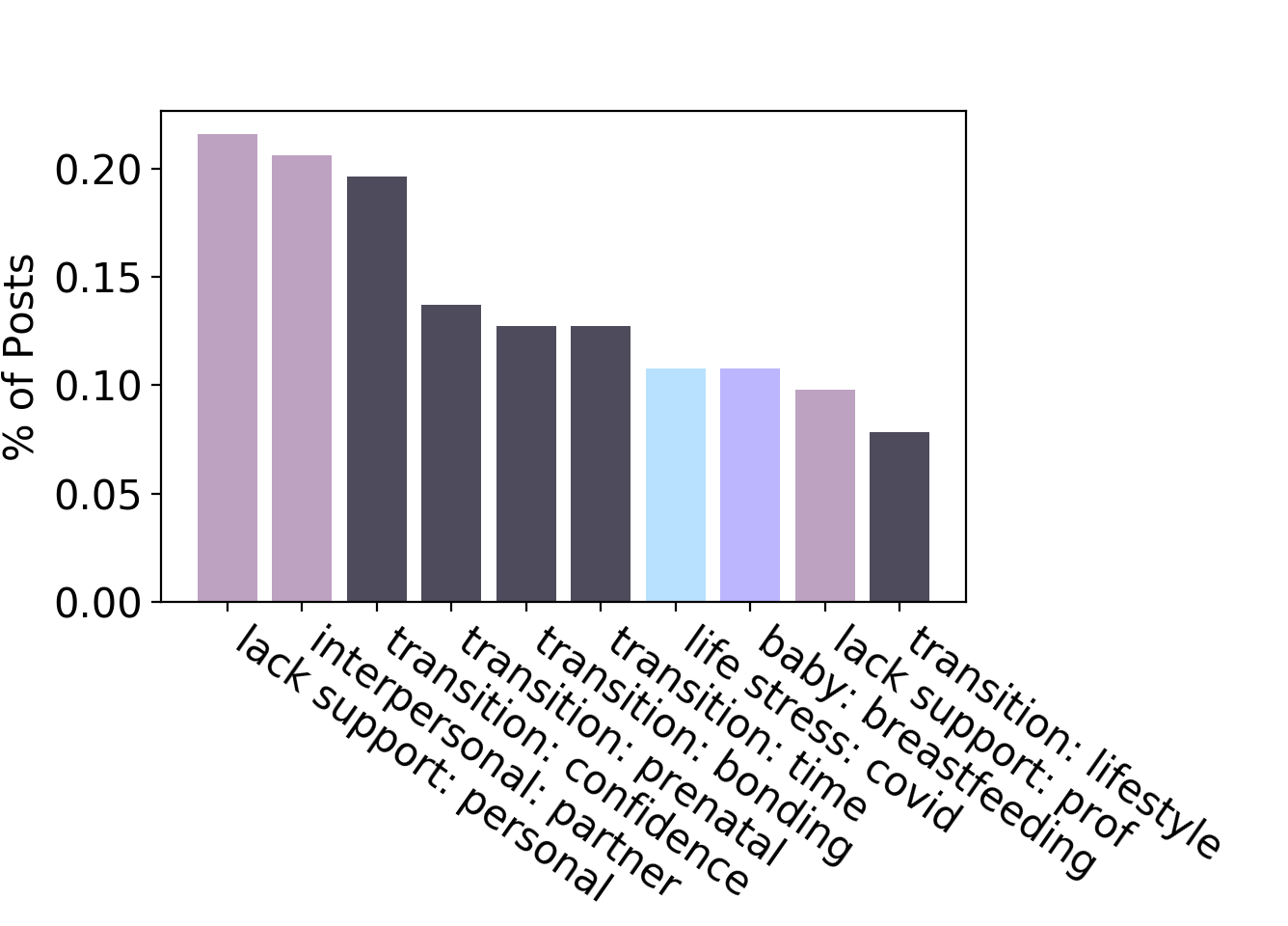} }}%
    \subfloat[\centering Concern categories of r/ppd]{{\includegraphics[width=0.5\linewidth]{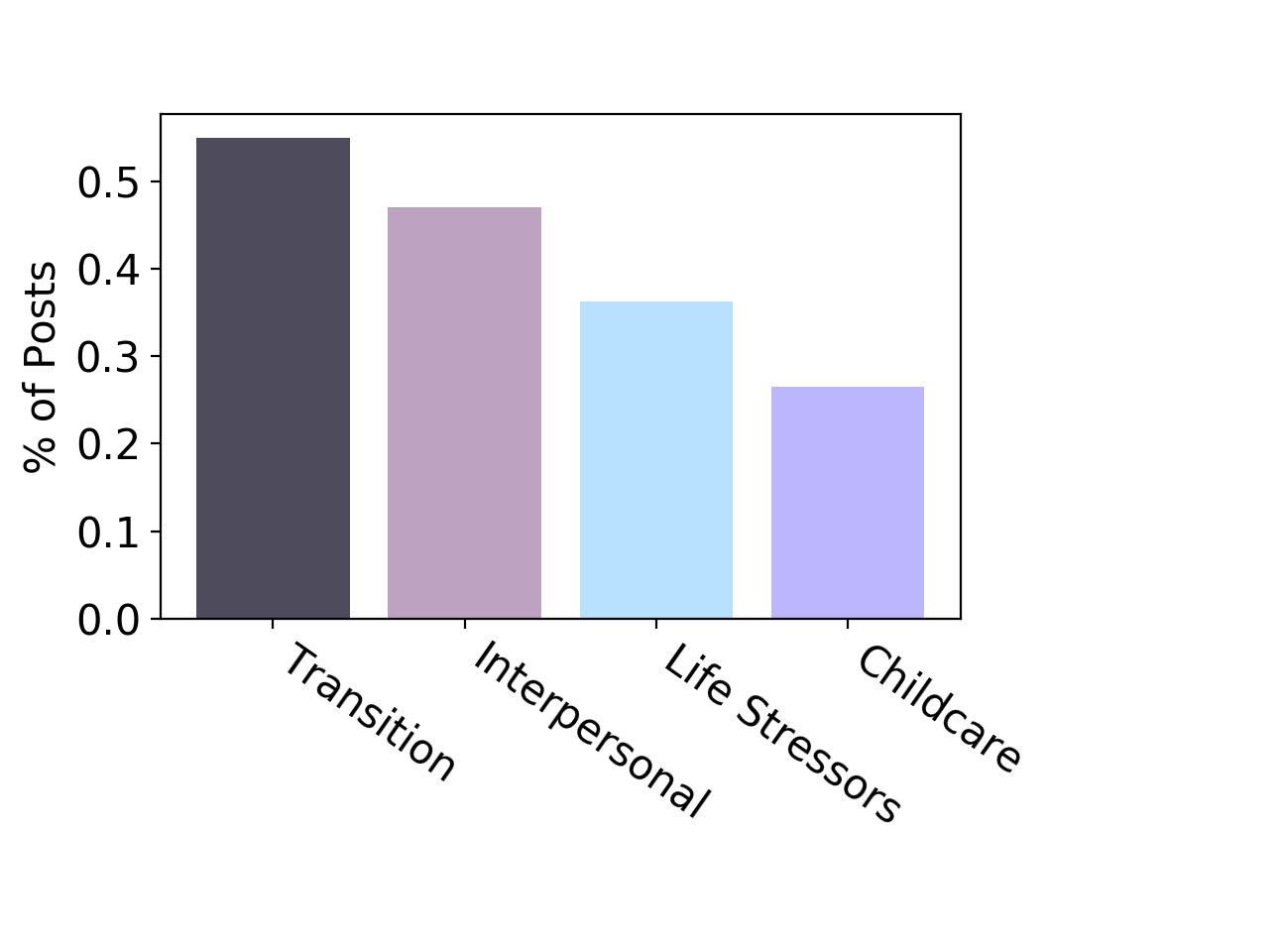} }} \\%
    \subfloat[\centering Individual concerns of r/NewParents]{{\includegraphics[width=0.5\linewidth]{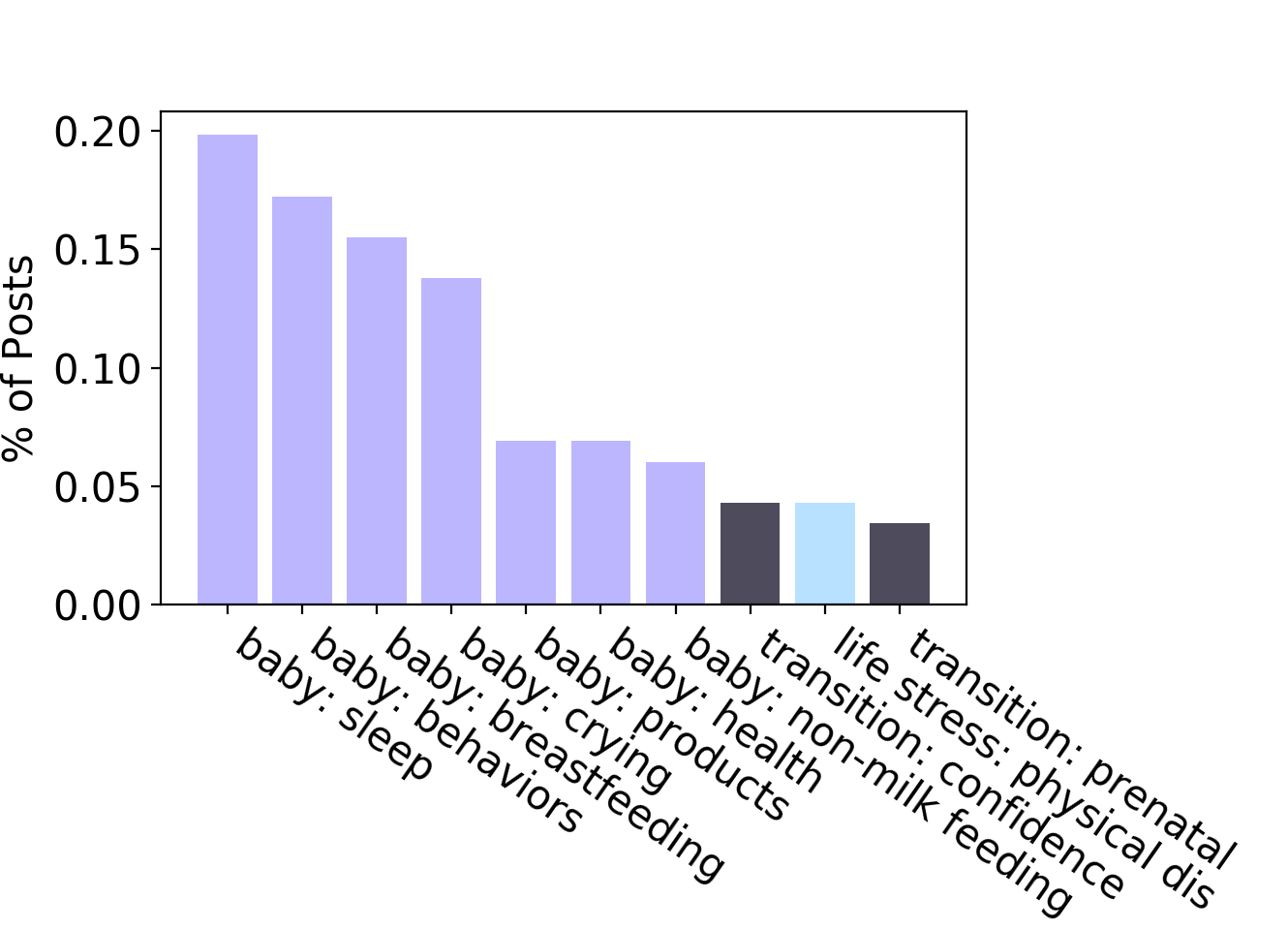} }}%
    \subfloat[\centering Concern categories of r/NewParents]{{\includegraphics[width=0.5\linewidth]{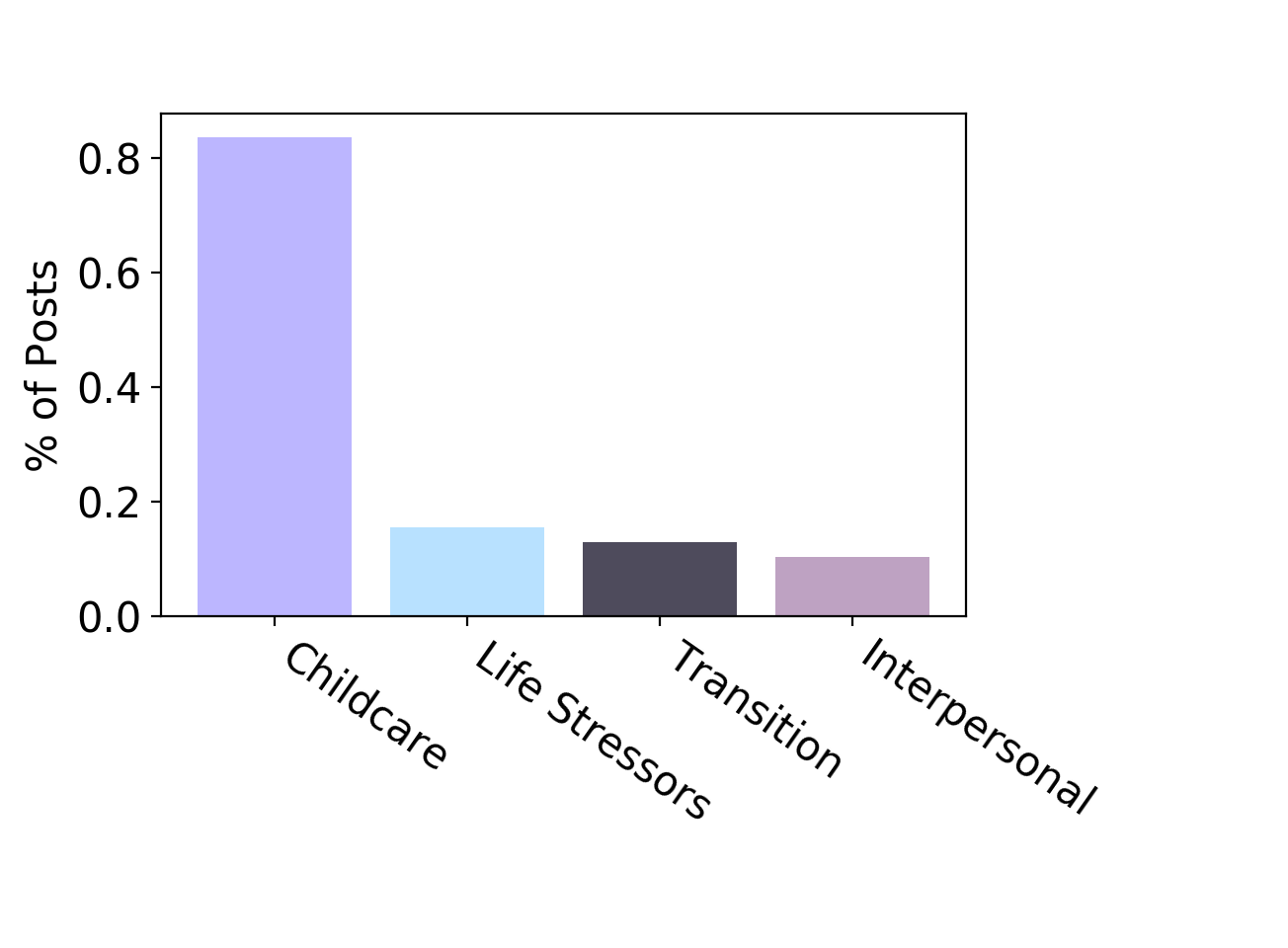} }}%
    \caption{Percentage of support seekers mentioning individual concerns and concerns as aggregated into categories from helpline and Reddit datasets as annotated by human coders.}%
    \Description{Bar charts showing top 10 individual concerns and concern categories for all 3 datasets.}
    \label{fig:annotation}%
\end{figure}

We utilized human annotation to analyze the concerns of postpartum support seekers in both platforms. Figure~\ref{fig:annotation} shows the top 10 individual concern topics mentioned by support seekers (left column), grouped by categories of concerns (right column). To highlight the relative proportion of concerns shared across different groups, we removed posts and conversations that did not mention any concerns from these plots, accounting for 52 conversations (12.78\%) from the helpline dataset. Major differences emerged between the concerns of mothers posting to the general parenting online forum (r/NewParents) and those posting to postpartum mental health platforms (helpline and r/ppd). In particular, over 80\% of "healthy" support seekers' concerns were focused on childcare, whereas "distressed" mothers' top two concerns were a perceived lack of support (24.51\% and 21.57\%) and interpersonal problems with their partner (20.85\% and 20.59\%). Furthermore, our annotation results indicate that 7 of the top 10 concerns for general parenting support seekers were in the childcare category, with the top four concerns being sleep (22.55\%), baby behaviors (19.61\%), breastfeeding (17.65\%) and crying (15.69\%). Typical sleep concerns included posts such as \textit{My 6 week old will not go back to sleep after I feed him at night.} and \textit{It's a struggle every night to get my 4 monther to sleep.} Typical posts about baby development, breastfeeding and crying included: \textit{My daughter is almost 10 months and still doesn't respond to her name.} and \textit{I might have to switch to formula if breastfeeding is such a problem.} and \textit{My first born was not nearly as fussy by this age}. 

"Distressed" mothers also occasionally expressed uncertainties or concerns about childcare. However, childcare concerns were substantially less frequent, with breastfeeding being the only childcare concern among the top 10 in both helpline and r/ppd (7.32\% and 10.78\%, ranked 7th and 8th respectively). Across platforms, mothers posting on postpartum mental health platforms were much more likely to describe difficulties beyond the inherent stresses of being a new parent per se. The most common category of concerns in both helpline and r/ppd posts was a lack of personal support. Several examples that illustrate the experiences of mothers with this concern include: \textit{I don't have any friends I can talk to about my ppd and my mom just keeps saying that every new mom feels like this.} and \textit{I'm feeling more and more depressed and somehow nobody seems to care.} These emotionally-laden descriptions of support seekers' experiences were frequent and often accompanied with other concerns that were exacerbated by their lack of support.

Moreover, drawing from our qualitative observations, we noted that when interpersonal problems were mentioned in r/NewParents, they were of a different nature than those mentioned by either of the r/ppd groups. Mothers from r/NewParents mostly blamed their partner for not being responsible enough with sentences like \textit{He won't help me with anything at night even though he's on leave from work. He just says that I can do it myself since I'm up and gets frustrated that I woke him up in the first place.} By contrast, interpersonal issues between "distressed" mothers and their partner revealed concerns about bonding and intimacy, with sentences such as \textit{He's home all day and just watches TV like he doesn't even care about his own son or his wife. I can't even tell if he loves me and he only talks to me when the baby needs to be changed or something.} and \textit{My husband and I were so close and cuddly and now he's always too busy or too tired and I just feel like he's not attracted to me anymore with the baby weight.}


In general, our annotation results indicate more commonalities than differences in the concerns shared by the postpartum communities on r/ppd and the helpline. However, we note that nearly twice as many postpartum mental health support seekers on Reddit indicated concerns in the transition to motherhood category relative to helpline support seekers. Mothers posting to r/ppd often mentioned challenges in their confidence as a mother and the weight of the numerous changes in their lives. For example, \textit{I'm not cut out to be a mom. I love my daughter and I'm trying to be the best mom I can, but I just look at her and think she deserves so much better than me.} and \textit{I'm stretched so thin taking care of my son. I can barely take a moment to eat or go to the bathroom or even just take a breath.}

\begin{figure}%
    \centering
    \subfloat[\centering   Helpline]{{\includegraphics[width=0.33\linewidth]{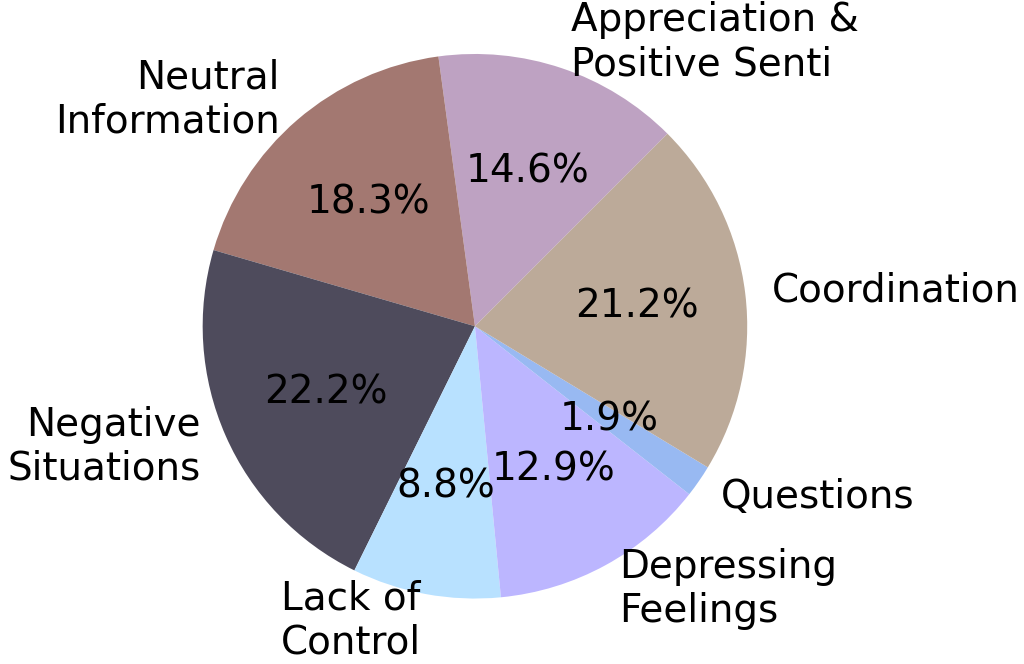} }}%
    \subfloat[\centering r/ppd]{{\includegraphics[width=0.35\linewidth]{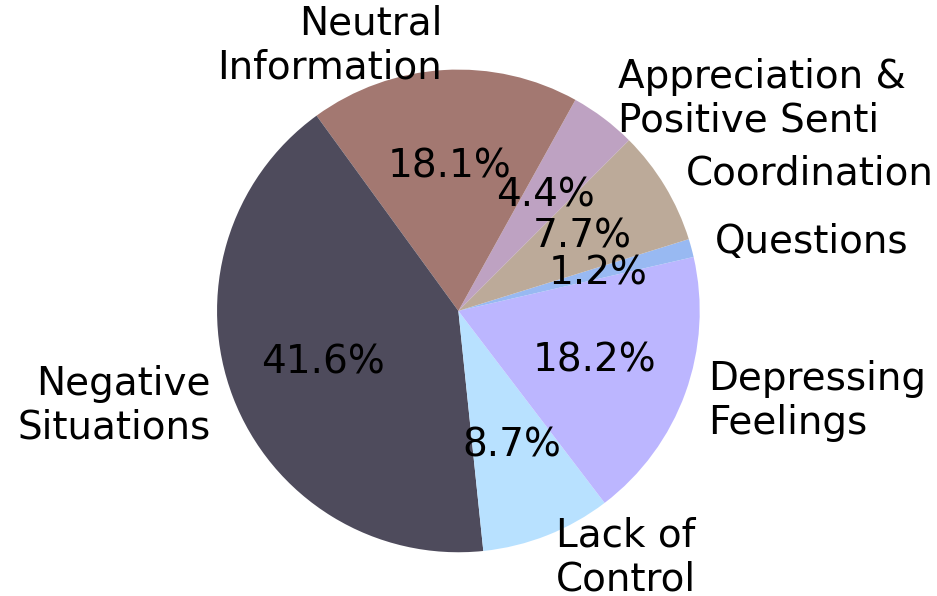} }}
    \subfloat[\centering r/NewParents]{{\includegraphics[width=0.31\linewidth]{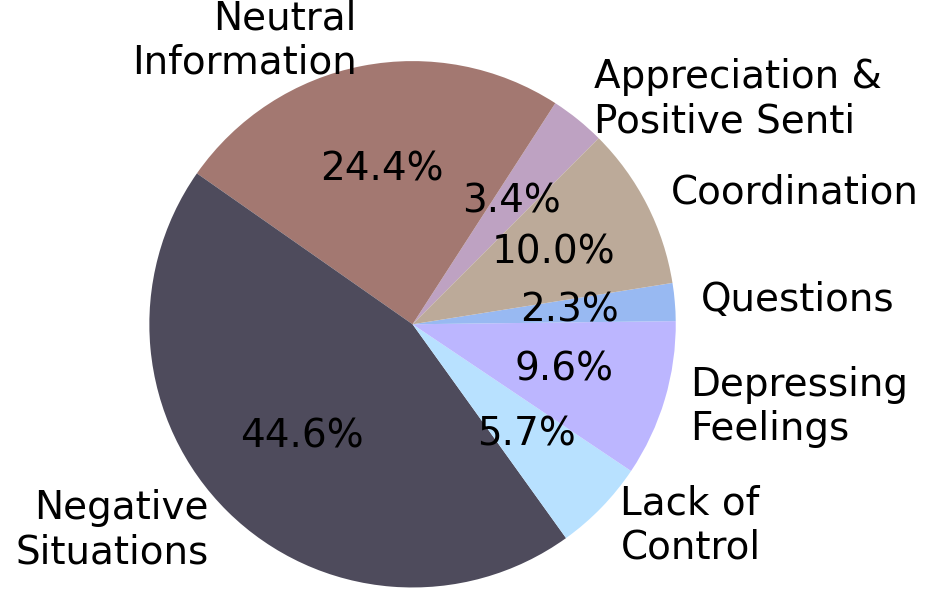} }} \\
    \caption{Percentage of sentences in each cluster for each dataset}%
    \Description{3 pie charts showing the percentage of sentences in each cluster for each dataset}
    \label{fig:pie}%
\end{figure}

These differences in the content of concerns were not reflected in the clustering or LIWC results. In particular, the clustering solution did not identify individual concerns or concern categories as shown in Figure~\ref{fig:pie} but instead provided a higher level categorization of sentences focused more on the affect and goals expressed by support seekers. Within the LIWC results, while the category "personal concerns" is ostensibly related to mothers' concerns (i.e. home, money, religion, and death), this list of concerns appears to be not specific enough to capture the concerns of postpartum mothers or new parents. Indeed, these words were rarely used by mothers sharing on the platforms in our dataset, and evidenced not much difference across groups (see Table~\ref{tab:liwc}).

 \begin{table}[!ht]
  \caption{Results of Psychological States from Human Coders}
  \label{tab:mental_state}
  \centering
  \begin{tabular}{cccc}
    \toprule
    \textbf{Psychological States} & \textbf{Helpline} &  \textbf{r/ppd} & \textbf{r/NewParents} \\
    \midrule
    Depressive Moods	&	64.62\%	&  76.70\%	& 18.80	\%	\\
    Anxiety	&	41.52\%	&	36.89\%	& 16.24\%	\\
    Severe symptoms	&	8.60\%	& 14.56\%	& 0		\\
    \bottomrule
  \end{tabular}
\end{table}


\subsection{What are the psychological states of support-seeking postpartum mothers?}

We leveraged annotations, dictionary-based text analysis, and unsupervised clustering to explore the psychological states of support seekers. 

Annotation results shown in Table~\ref{tab:mental_state} indicate worse psychological states in PSI helpline and r/ppd relative to r/NewParents across all annotated states, as expected. Mothers in the r/ppd group shared or evidenced depressive moods and severe symptoms at the highest rate. Mothers texting the helpline also shared high rates of depressive and severe symptoms. Our severe symptom annotation was comprised of suicidal thoughts, self harm, or other indications of safety concerns. The high rates indicate that mothers active on both postpartum mental health platforms were willing to share vivid descriptions of their pain and despair. For example, mothers wrote \textit{I can't help but wish that the doctor would have just let me die after the C-section complications} and \textit{I don't know how much longer I can keep going like this, hating myself so deeply and feeling so hopeless.} These high levels of despair point to the importance of these platforms as an outlet for mothers, as well as the potential for providing timely support at their darkest moments. 



LIWC results mirrored the annotations findings (see Table~\ref{tab:liwc}).  Consistent with prior research on patterns of speech in clinically depressed samples \citep{fbpredict, munmun_twitter2, predictingtwitter, depression_pasttense}, both helpline and r/ppd mothers used more 1\textsuperscript{st} person singular ("I"), fewer 3\textsuperscript{rd} person singular ("he/she"), and fewer articles (e.g. "the", "an") than r/NewParents mothers. Besides, both postpartum mental health groups used more negation, more negative emotion words, and wrote more frequently in the past tense relative to support seekers posting to the general new parents forum. Similar to what we found with annotations, support seekers from r/ppd evidenced more depressive symptoms relative to helpline support seekers. They used more negative emotion words, specifically in the anger and sadness word categories, and fewer positive emotion words. Surprisingly, helpline support seekers used the most positive emotion words despite their relatively high amount of negative emotion words. We suspect this might be because of the emotional validation and encouragements offered by helpline volunteers. Additionally, relative to r/ppd and r/NewParents, helpline support seekers used more 2\textsuperscript{nd} person pronouns ("you"), likely owing to the conversational nature of the helpline platform. 

\begin{figure}[h]
  \centering
  \includegraphics[width=\linewidth]{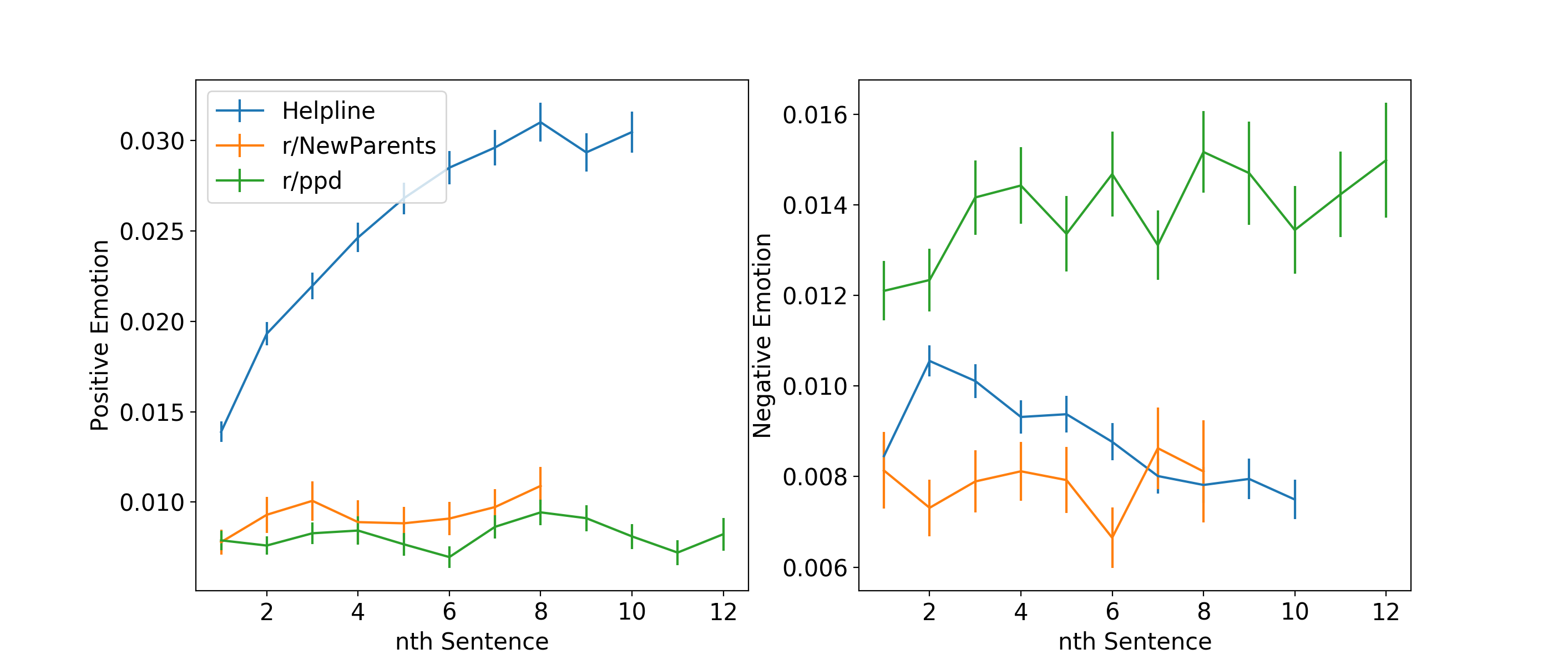}
  \caption{Sentence-by-sentence positive and negative emotion calculated over each dataset.}
  \label{fig:sentiment}
  \Description{2 line charts showing sentence-by-sentence positive and negative LIWC emotion change calculated over each dataset.}
\end{figure}


\begin{table}[!ht]
  \caption{LIWC Linguistic Text Comparisons}
  \label{tab:liwc}
  \centering
  \begin{threeparttable}
  \begin{tabular}{lcccccc}
    \toprule
    \multirow{2}{*}{\textbf{Category}} & \multicolumn{2}{c}{\textbf{Helpline}}  & \multicolumn{2}{c}{\textbf{r/ppd}} & \multicolumn{2}{c}{\textbf{r/NewParents}}  \\\cmidrule(l{0.5em}r{0.5em}){2-3} \cmidrule(l{0.5em}r{0.5em}){4-5}\cmidrule(l{0.5em}r{0.5em}){6-7}
    & mean & std & mean & std & mean & std \\
    \midrule
    1\textsuperscript{st} Person Singular	&	\textbf{0.035}* & 0.011 & \textbf{0.034}+ & 0.012 & \textbf{0.017}\# & 0.011 \\
    1\textsuperscript{st} Person Plural		& 0.001* & 0.002 & 0.002+ & 0.003 & 0.004\# & 0.006 \\
    2\textsuperscript{nd} Person Pronoun	& 0.004* & 0.006 & 0.001 & 0.004 & 0.002\# & 0.003 \\
    3\textsuperscript{rd} Person Singular	& \textbf{0.005}* & 0.007 & \textbf{0.008}+ & 0.008 & \textbf{0.014}\# & 0.01 \\
    3\textsuperscript{rd} Person Plural		& 0.001 & 0.002 & 0.001 & 0.003 & 0.002\# & 0.004 \\
    \hline
    Articles	&	\textbf{0.012}* & 0.007 & \textbf{0.012}+ & 0.006 & \textbf{0.015}\# & 0.008 \\
    \hline
    Positive emotion & 0.013* & 0.01 & 0.008 & 0.005 & 0.008\# & 0.007 \\
    Negative emotion	& \textbf{0.009}* & 0.007 & \textbf{0.012}+ & 0.007 & \textbf{0.007}\# & 0.007 \\
    \hspace{3mm} Anxiety	& 0.003 & 0.004 & 0.003+ & 0.004 & 0.002\# & 0.004 \\
    \hspace{3mm} Anger	 & 0.001* & 0.002 & 0.002+ & 0.003 & 0.001\# & 0.003 \\
    \hspace{3mm} Sadness	& 0.004* & 0.004 & 0.004+ & 0.004 & 0.002\# & 0.003 \\
    \hline
    Past focus	 & \textbf{0.012}* & 0.008 & \textbf{0.013}+ & 0.008 & \textbf{0.01}\# & 0.008 \\
    Present focus	& 0.048* & 0.011 & 0.038+ & 0.013 & 0.036\# & 0.012 \\
    Future focus	& 0.004* & 0.004 & 0.003 & 0.003 & 0.003\# & 0.004 \\
    \hline
    Negation & \textbf{0.008}* & 0.005 & \textbf{0.006}+ & 0.005 & \textbf{0.004}\# & 0.004 \\
    Swear words	& 0.0* & 0.001 & 0.001 & 0.001 & 0.001\# & 0.002 \\
    Nonfluencies & 0.001* & 0.002 & 0.0 & 0.001 & 0.0\# & 0.002 \\
    \hline
    Work & 0.004* & 0.005 & 0.003 & 0.004 & 0.003\# & 0.004 \\
    Leisure & 0.001* & 0.002 & 0.001+ & 0.002 & 0.002\# & 0.004 \\
    Home & 0.001* & 0.002 & 0.002 & 0.002 & 0.002\# & 0.004 \\
    Money & 0.001* & 0.002 & 0.001 & 0.001 & 0.001 & 0.002 \\
    Religion & 0.0 & 0.001 & 0.0 & 0.001 & 0.0 & 0.001 \\
    Death & 0.0* & 0.001 & 0.001+ & 0.007 & 0.0 & 0.001 \\

    \bottomrule
  \end{tabular}
  \begin{tablenotes}\footnotesize
    \item {\raggedleft \small   * denotes significant difference between helpline and r/ppd \par}
    \item {\raggedleft \small   + denotes significant difference between r/ppd and r/NewParents \par}
    \item {\raggedleft \small   \# denotes significant difference between helpline and r/NewParents \par}
    
   \end{tablenotes}
  \end{threeparttable}
\end{table}

To examine the temporal dynamics of mothers' affective experiences on the platforms themselves, we examined sentence-by-sentence changes in LIWC positive and negative emotion word categories. Figure~\ref{fig:sentiment} shows the sentence-by-sentence changes in the percentage of negative and positive words in consecutive sentences for each dataset. We averaged proportions across each respective dataset and calculated standard deviation. We included data up to the median number of sentences per dataset to avoid spurious variations associated with low N. Results indicate that in both "healthy" and "distressed" Reddit datasets, positive and negative emotions were stable across sentences within a post. However, in the helpline dataset, positive emotions increased and negative emotions decreased through the conversation. In the discussion, we consider how differences between the two platform features may have shaped these trajectories of affect while also constraining our ability to document additional possible affect changes in the Reddit data.


Our clustering solutions appeared to highlight differences in the emotional tone of texts shared by mothers' that were also related to postpartum mental health status. In particular, two of seven clusters were characterized by descriptions of negative situations and descriptions of depressing feelings. All three datasets have large negative situation clusters. In fact, the r/NewParents has the highest proportion of negative situation sentences (44.6\% vs 22.2\% and 41.6\% from helpline and r/ppd). However, the size of the negative situation cluster does not speak to the degree or severity of shared challenges. In particular, each of the three platforms are used by caregivers seeking support and advice, and thus it is not unexpected that caregivers across these platforms often provide descriptions of challenging situations. 

The ratio of sentences in the negative situation cluster relative to the depressive feelings cluster may more accurately represent the emotional tone of mothers' descriptions of their challenges. Using this statistic, r/NewParents posts have substantially higher ratio of sentences in the negative situation vs. depressive feelings clusters (4.65) relative to the helpline and r/ppd (1.72 and 2.29), reflecting the higher depressive mood states of "distressed" mothers. Additionally, there were significantly more sentences in the appreciation and positive sentiment cluster within helpline (14.6\%) compared to Reddit groups (4.4\% and 3.4\%), consistent with results from LIWC.


\subsection{What are the goals of postpartum support seekers?} \label{sec: motivations}
Our annotations provide insight into both implicit and explicitly stated goals of support seekers contacting digital platforms as shown in Table~\ref{tab:specific_request}. 

We found that general support and venting or talking with someone were the top two annotated goals across both postpartum mental health platforms and the general parenting platform. However, the nature of those goals differed greatly. For helpline and r/ppd, support seekers were more personal and asked about depressing situations and feelings, with sentences such as \textit{Is it normal to have these kinds of thoughts and extreme emotions?} and  \textit{Just making it through the day is a struggle at this point and I need some advice on how to make it a little easier.} On the contrary, general support requested from r/NewParents matched top concerns in Section~\ref{sec: topics} which were dominated by seeking information on childcare-related issues with sentences like \textit{Any tips on how to have a smooth transition from breastfeeding to formula?} and \textit{Does teething make your baby hungrier?} For talking with someone, we observed that support seekers from r/NewParents tried to "get it off their chest" and sought a feeling of relief with sentences like \textit{I'm pissed off and just want to vent.} and \textit{Not a huge deal, but very very annoying.} while utterances from helpline data reflected more distress and matched the top concern in Section~\ref{sec: topics}--lack of support. Example sentences include \textit{I really need someone to talk to right now or I'm gonna have a breakdown.} and \textit{I'm not sure what I'm looking for, but I needed to let that out somewhere. Maybe there's someone who might get what I'm feeling.}

Our clustering results also indicated substantial differences in the resource-related content shared on the two platforms. In particular, over one fifth of the sentences (21.2\%) from helpline support seekers were about resource coordination, relative to 7.7\% and 10.0\% in the r/ppd and r/NewParents groups, respectively (see Figure~\ref{fig:pie}). We consider the implications of these platform-related differences further in the discussion. Annotation results also indicated that helpline support seekers requested diverse sources of support, with 14.49\% of the support seekers asking about therapist/support group information, 7.86\% asking about resources based on specific constraints such as finances or insurance, and 5.16\% asking about local coordinators in their area. 


\begin{table}[!h]
  \caption{Results of Goals from Human Coders}
  \label{tab:specific_request}
  \begin{tabular}{cccc}
    \toprule
    \textbf{Goals} & \textbf{Helpline} & \textbf{r/ppd} & \textbf{r/NewParents} \\
    \midrule
    Local Coordinator & 5.16\% & 0 & 0 \\
    Talk With Someone & 34.40\% & 24.27\% & 10.26\% \\
    Health Care Provider & 14.49\% & 0.97\% & 0.0 \\
    General Support & 47.91\% & 51.46\% & 87.18\% \\
    Resource Barrier & 7.86\% & 0.0 & 0.0 \\
    PSI Information & 15.72\% & 0.0 & 0.0 \\
     
  \bottomrule
\end{tabular}
\end{table}

\subsection{Concerns, states and goals of support seekers other than mothers} \label{sec: others}
Our sample of annotated non-maternal postpartum support seekers consisted of 35 fathers, 9 health care providers, 13 friends or family members, and 5 people with an unknown relationship to the mother. Of the 35 fathers, two (one from the helpline, one from r/ppd) were seeking support for themselves. The rest of the fathers (94\%) were asking for resources and advice on how to best support their partners who were experiencing postpartum distress. These digital platforms are not meant to be exclusive to postpartum mothers, but at this time, it appears that they are primarily used by mothers and those trying to support a postpartum mother.

\subsection{What are the contributions of machine analyses versus human annotations in understanding support seekers' concerns?} \label{sec: compute_human}

\begin{figure}%
    \centering
    \includegraphics[width=1\linewidth]{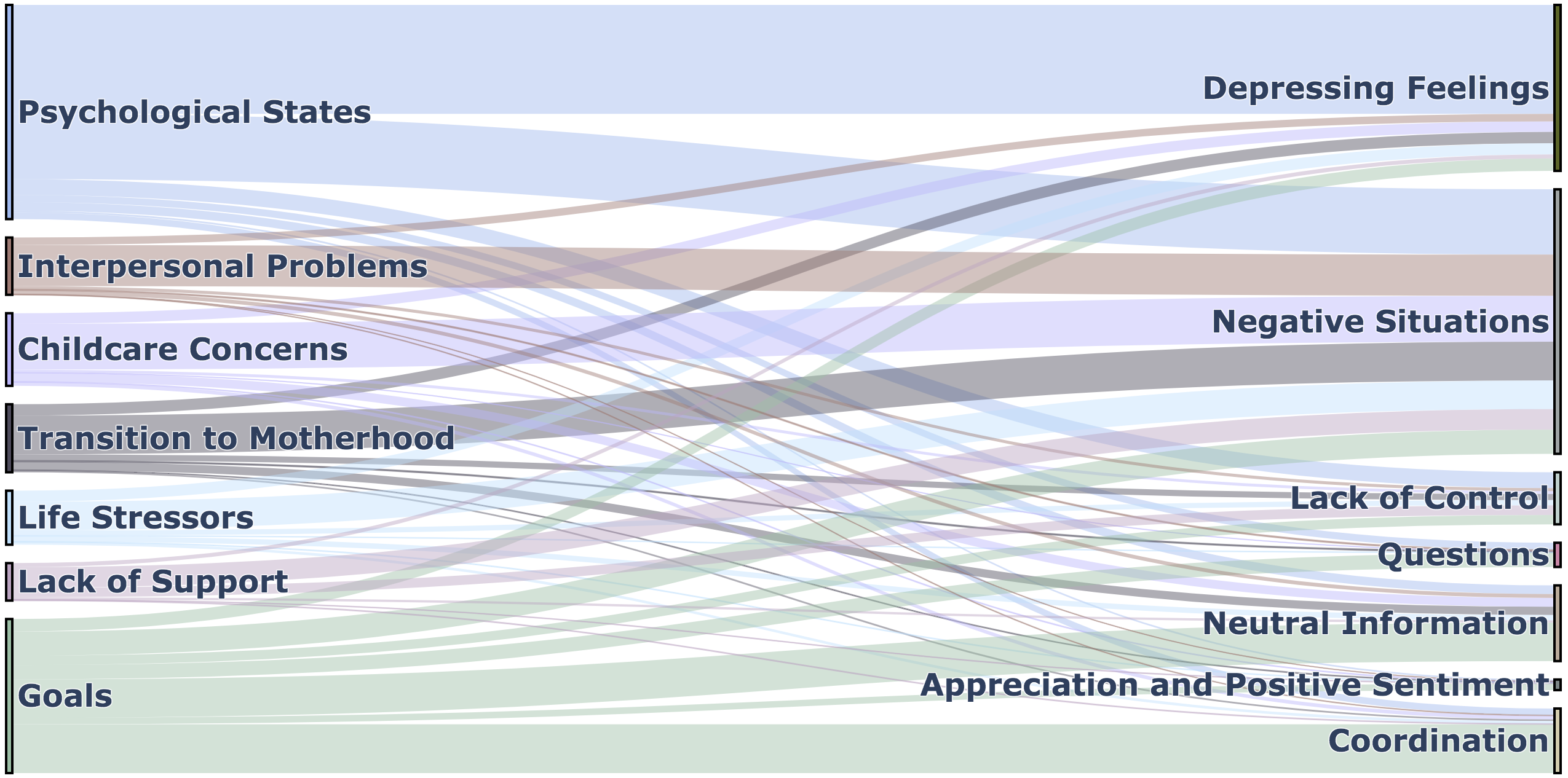} 
    \caption{Sankey diagram showing agreement between human annotation and unsupervised clustering.}%
    \Description{Sankey diagram showing agreement between human annotation and unsupervised clustering.}
    \label{fig:sankey}%
\end{figure}

To systematically compare the contributions of the annotations and the unsupervised clustering results, we match each annotated sentence with its assigned cluster as shown in Figure~\ref{fig:sankey}. This plot includes all annotated sentences across all three datasets. Overall, we note that the results from clustering align with manual annotations. For example, 50.85\% and 30.49\% of the sentences annotated as "psychological states" are clustered as "depressing feelings" and "negative situations", respectively. The majority of sentences annotated as concerns (71.81\% of interpersonal problems, 63.23\% of childcare concerns, 57.06\% of transition, 53.02\% of life stressors, and 54.87\% of lack of support) are matched to the "negative situations" cluster. 31.67\% and 24.94\% of sentences annotated as indicating mothers' "goals" are clustered as "coordination" and "neutral information". However, 15.71\% of the sentences of "goals" are clustered into "negative situations". A quick inspection reveals these sentences contain "negative emotions" words such as the phrase "postpartum depression", in contexts where human annotators would not indicate true negative emotion, such as \textit{I'm looking for a group or any resources for postpartum depression/anxiety.} To summarize, inspecting clusters provides us with a reasonably accurate indication of the high-level content shared by support seekers, which both overlaps with and complements the human annotations.

\section{Discussion}
Digital platforms have emerged as avenues of support for caregivers during the postpartum period-- a time marked by significant change and complex emotional experiences. Our descriptive analyses provide insight into the experiences of postpartum support seekers - their challenges, psychological states, and goals for reaching out. Using a variety of complementary analyses, we found a number of key differences between content shared by mothers posting on postpartum mental heath platforms and more general parenting platforms, between content shared on online forums and digital helplines for postpartum distress, and in the nature of results obtained from the various techniques we used. We summarize these differences below.


\subsection{Mothers' posts on postpartum distress vs general parenting platforms}

\paragraph{Concerns}
The majority of mothers posting to r/NewParents, a general parenting platform, focused on childcare issues such as sleep and behaviors, consistent with previous work \cite{diyi_onlineMedia}. 
By contrast, mothers posting to r/ppd and the PSI helpline focused on lack of support, interpersonal problems, and transition to motherhood. It is well-established that social support serves as a buffer for postpartum mood disorders \cite{socialsupport_buffer}, and is in line with our findings of mothers reporting a lack of personal or professional support more frequently on the postpartum mental health platforms. Questions regarding childcare were also present on the postpartum mental health platforms, reflecting the ubiquity of barriers to information among new parents and the ways in which digital platforms can bridge these barriers. 


\paragraph{Psychological States}
Mothers described severe symptoms, such as suicidal ideation, in 8.60\% of helpline posts and 14.56\% of r/ppd posts, compared with 0 mentions in r/NewParents. Postpartum mood disorders can lead to increased feelings of guilt and worthlessness \cite{postpartum_guilt}, which can underlie challenges in the transition to motherhood like parenting confidence--a common concern on postpartum mental health platforms. Furthermore, the interpersonal problems and life stressors that mothers described on the postpartum mental health platforms (e.g., infidelity or job-loss), can also trigger the onset of a mood disorder or exacerbate an existing one \cite{interpersonal_depression}. While our data cannot distinguish between cause and effects, many mothers reaching out to PSI helpline and r/ppd were experiencing serious emotional distress that we did not see in the r/NewParents posts.



\paragraph{Goals}
Mothers had more frequent goals of talking with someone when using helpline and r/ppd compared to r/NewParents. The high rate of lack of support and frequent descriptions of negative situations and depressive emotions implies that mothers seem to turn to online forums and peer-to-peer support after they have exhausted their natural supports. They hope that they can get better advice and validation than what they have access to currently.

Additionally, a considerable number of mothers utilizing postpartum mental health platforms expressed concerns about going to an emergency department or telling a medical professional about their suicidal thoughts, in fear that their baby would be taken from them. As such, we can infer that one important function of such platforms is that they are perceived by mothers as safer alternatives to disclose the severity of their distressing thoughts and emotions. Assessing the potential for actual harm and linking mothers in need of emergency services is a key need for these platforms, as is providing kind and reassuring support to mothers who are distressed by these feelings  \cite{ppd_obsessions}.


\subsection{Online Forum vs Digital Helpline for Postpartum Distress}
We observed both similarities and differences between distressed caregivers content shared on PSI and the r/ppd forums. 

Distressed support seekers on both platforms indicated that a lack of support and interpersonal problems were major issues for them. However, we also observed some differences in content shared on PSI helpline vs. Reddit. Distressed support seekers on Reddit shared more depressive moods (76.70\% vs. 64.62\%) and severe symptoms (14.56\% vs. 8.60\%) relative to support seekers on PSI. Distressed reddit support seekers were also more likely to share concerns around their transition to motherhood, including challenges bonding with their baby and a sense of loss of their pre-birth identies. These additional disclosures may be due to the complete anonymity of online forums, where users view it as more acceptable to share potentially stigmatizing content \cite{virtual_voices}. The differences in the rates of severe symptoms suggests that support seekers may have fears of being reported when talking to trained volunteers. Platforms may be able to increase disclosure of stigmatizing content by indicating whether they are mandatory reporters of such information. 

Our results also indicate a platform-specific effect of mood changes across shared content. Specifically,  mood states identified by LIWC remained consistent across sentences of posts shared on both Reddit platforms, but positive affect words increased and negative affect words decreased across sentences shared on digital helpline conversations. In one sense, this is an "unfair" comparison, given that digital helplines are turn-based platforms and Reddit is not.  However, the turn-taking nature of digital helplines is an essential distinguishing feature of such platforms that should be considered for its potential to offer a very different experience of support. Our data suggests that turn-taking conversations afforded by the helpline platform - or the real-time validation, encouragement, or resources shared by the PSI volunteers - worked to shift helpline support seekers expressed mood over their conversations. We note that support seekers posting on Reddit may feel better after reading responses to their post. However, as noted in the introduction, many posts do not receive any comments, and it is even more rare for the original support seeker to reply. Relatedly, our LIWC results indicated that helpline support seekers used the most 2\textsuperscript{nd} person pronouns ("you"), positive emotion words, present and future tense relative to support seekers on both subreddits. This does not reflect the typical patterns of depressed speech: rather, it likely captures another way in which the conversational nature of the helpline platform changes what support seekers write and feel. Thus, our results suggest that platforms designed for real-time conversational "feedback" are likely to lead to immediate increases in support seekers affect. However, given that we do not have a measure of potential changes in affect over hours or days, our work cannot speak to the potential longer-term affect trajectories, specifically in the case of online forums such as Reddit where responses are received over longer periods of time.

Finally, our clustering results also showed that the coordination cluster was 2-3 times larger for the helpline than either of the two Reddit platforms, which indicate directly that helpline volunteers are responding to the requests for information, in their attempts to provide support seekers with local resources. However, responses posted on Reddit may also contain concrete resources, which we did not have access to in our analyses. Platforms which provide specific resources and support can be an important source of support for caregivers experiencing postpartum distress. Concrete virtual or local resources can provide critical support for these caregivers and our results suggest that mothers are engaging with PSI volunteers to obtain these resources.



\subsection{Human Annotation vs. Unsupervised Clustering}

Each of the three analysis techniques provided unique yet complementing insights into postpartum support seekers' experiences. The advantage of human annotations is that domain expertise can focus the analyses on the particular questions and distinctions of greatest interest to the researchers. Our annotations provided rich qualitative insights into the concerns, emotional states, and goals of support seekers. However, manual annotation is especially laborious in that it can take considerable time to design annotation schemes, reach inter-rater agreement, and label the dataset.

On the other hand, the dictionary-based and unsupervised learning techniques we used did not require the use of manual labelling, meaning that results could be obtained within a short period of time. This allows for quick experimentation and testing of hypotheses. However, results may not be of strong interest to researchers as they may indicate more generic structure in the data. Additionally, the hyperparameters of unsupervised clustering have a huge impact on the results and hyperparameter tuning can be troublesome, which makes this technique unreliable compared to manual annotation. We found that the dictionary-based technique LIWC effectively captured the affect at the scale of the entire dataset and also in fine temporal detail, replicating past results about affect of mothers with postpartum distress while also adding to this work by showing stability and change across helpline vs. Reddit support seekers affect across their shared content. While our clustering results did not reveal distinguishing concerns of "distressed" versus "healthy" caregivers, they appeared to capture differences in the types of information shared by support seekers, highlighting differences in the quality and intensity of affect mothers used to describe their situations (neutral information vs. negative situations; negative situations vs. depressing feelings) and whether mothers were sharing vs. asking for information. 

Comparing the results across all three techniques, we found that our human annotations indicated the most meaningful differences between the distressed and non-distressed caregivers. In particular, human annotations precisely indicated the types of concerns that are both shared and unshared between caregivers with and without postpartum distress. As such, human annotations rather than clustering or LIWC provided a clearer answer to our main research questions regarding the unique needs of caregivers experiencing postpartum distress. However, given that clustering and LIWC results appear to broadly indicate the affect and types of information shared by support seekers, as indicated in Section~\ref{sec: compute_human}, we can leverage such techniques to support the development of tools to support caregivers experiencing postpartum distress. In ongoing work, we are developing a chatbot that can respond to common concerns of postpartum support seekers. Our current efforts parse each input obtained by a support seeker by combining supervised algorithms trained using our human annotations with sentence cluster information to provide both precise information about the content of the input as well as the high-level context of support seekers’ conversational needs or goals. In this way we leverage the complementary strengths of both human annotations and unsupervised techniques.

\subsection{Other Support Seekers}

Mothers are not the exclusive users of the postpartum platforms. Each platform had a small percentage of posts from fathers, family members, and social workers reaching out to inquire about resources and advice for someone who is experiencing postpartum distress. While postpartum mood disorders affect fathers too, our analyses revealed only two posts from fathers seeking support for themselves, rather than their partners. As such, we did not have enough data to speak to the specific postpartum concerns of fathers. Prior work suggests that gender roles and stigma affect men's willingness to reach out for help \cite{male_help}, demonstrating how the impact of sexism and patriarchal systems can extend beyond its obvious negative effects on mothers. In particular, the lack of support seeking and the complex systems maintaining this can keep fathers from addressing their postpartum emotional experiences, with negative implications for themselves and their ability to support mothers. Given the frequency of mothers' sharing about interpersonal concerns with partners and a lack of support in our helpline and r/ppd datasets, the well-being of fathers in the postpartum period may be an essential aspect of increasing social support for mothers.

\subsection{Limitations and future work}
As we examined only one online forum and one digital helpline in our analyses, our work does not reflect the diversity of support seeking behavior that exists across different online forums or digital helplines. For example, while Reddit posts are generally anonymous and comments are open to the public, users on Facebook can choose their targeted audience and are often known by their readers which could influence the content they share. Future research could examine the differences within and between these two digital platforms to better understand and provide support. Additionally, our sample was largely English-speaking support seekers from the US, whose experiences may not be generalizable to the broader postpartum population. Other limitations come from the small size of our datasets in comparison with previous research \citep{diyi_onlineMedia, reddit_nlp_covid}. While the samples we collected and annotated are completely random, we acknowledge that potential biases may be introduced in manual annotation. Additionally, concerns such as interpersonal problems regarding violence and life stressors about medication were discarded because of their infrequency or low kappa among annotators, which makes our results incomplete. Future studies could leverage our annotated dataset to develop supervised models that could be used to analyze larger volumes of data, either independently or as a complement to the dataset of \citep{diyi_onlineMedia}. Lastly, because "distressed" support seekers expressed more concern about poor social support and partner problems relative to general parenting questions, future work should evaluate the efficacy of targeted educational materials designed to directly address these concerns.

\section{Conclusion}
Our paper leverages human insights and data-oriented methods to provide a descriptive analysis of the experiences of postpartum seekers on two digital platforms. We found major differences in the concerns and psychological states of "healthy" and "distressed" mothers. We found meaningful differences in the patterns of speech and affect of mothers sharing their experiences on a postpartum mental health support helpline vs. online forum. And we found that human-derived annotations and unsupervised clustering provided distinct yet overlapping insights into the content shared by postpartum support seekers. In summary, we aim for this work to bring attention to the postpartum community impacted PPMADs in order to provide effective understanding and support via digital platforms. Our analyses have revealed specific needs for validation as many support seekers described a profound lack of support both at home and with professionals. Fathers also need more encouragement to reach out as we observed few such occurrences on both platforms, even when anonymity was guaranteed. We further hope the insights from this work could lead to the greater use and development of natural language processing techniques to provide better and more robust tools for postpartum parents.








\paragraph{Annotation Scheme}
We developed the annotation scheme with the goal of comprehensively categorizing support seekers' concerns, their reported psychological symptoms and the types of requested support. After reading and discussing the prevailing themes of concerns found in 100 of the longest conversations from the helpline data set, five research assistants, including two of the authors drafted an initial annotation scheme with items divided into categories associated with support-seeker concerns, psychological states, and goals. Support-seeker concerns included five subcategories, i.e. interpersonal problems, childcare concerns, life stressors, transition to motherhood, and lack of support.


\begin{acks}
This work was supported by the National Institute of Mental Health K01 Award (1K01MH111957-01A1) to Kaya de Barbaro and Good Systems, a research grand challenge at the University of Texas at Austin. We thank support seekers from PSI and Reddit for helping us get a better understanding of the experience of postpartum women and design a better version of one-to-one text platform.
\end{acks}





\bibliographystyle{ACM-Reference-Format}
\bibliography{sample-base}


\end{document}